\documentclass[runningheads]{llncs}
\usepackage{graphicx}
\usepackage{amsmath}
\usepackage{amssymb}
\usepackage{booktabs}

\usepackage[thinlines]{easytable}
\usepackage{enumitem}
\usepackage{multirow}
\usepackage{tabu}
\usepackage{xcolor}
\usepackage{transparent}

\usepackage{tabulary}
\usepackage{amsfonts}       % blackboard math symbols
\usepackage{nicefrac}       % compact symbols for 1/2, etc.
\usepackage{microtype}      % microtypography
\usepackage{enumitem}
\usepackage{tabularx}
\usepackage{adjustbox}
\usepackage{subfig}
\usepackage{lipsum}
\usepackage{scalerel}

\newcommand{\figref}[1]{Fig.~\ref{#1}}
\newcommand{\tabref}[1]{Table~\ref{#1}}

\newcommand{\secref}[1]{Sec.~\ref{#1}}
\newcommand{\etal}{{\it et al.}}

\usepackage{tikz}
\usepackage{comment}
\usepackage{amsmath,amssymb} % define this before the line numbering.
\usepackage{color}

\usepackage[accsupp]{axessibility}  % Improves PDF readability for those with disabilities.

\usepackage[pagebackref=true,breaklinks=true,letterpaper=true,colorlinks,bookmarks=false]{hyperref}
\hypersetup{
     colorlinks   = true,
     citecolor    = green
}
\hypersetup{linkcolor=red}

\begin{document}

\newcommand\blfootnote[1]{%
  \begingroup
  \renewcommand\thefootnote{}\footnote{#1}%
  \addtocounter{footnote}{-1}%
  \endgroup
}

% \linenumbers
\pagestyle{headings}
\mainmatter
\def\ECCVSubNumber{5441}  % Insert your submission number here

\title{
Bridging Images and Videos: \\
A Simple Learning Framework for Large Vocabulary Video Object Detection}
% A Unified Learning Framework for \\ Large Vocabulary Video Object Detection}

% \begin{comment}
% \titlerunning{ECCV-22 submission ID \ECCVSubNumber} 
% \authorrunning{ECCV-22 submission ID \ECCVSubNumber} 
% \author{Anonymous ECCV submission}
% \institute{Paper ID \ECCVSubNumber}
% \end{comment}
%******************

% \begin{comment}
\titlerunning{A Simple Learning Framework for Large Vocabulary Video Trackers}
\author{
Sanghyun Woo$^{1*}$ \and
Kwanyong Park$^{1*}$ \and \\
Seoung Wug Oh\inst{2} \and 
In So Kweon\inst{1} \and
Joon-Young Lee\inst{2}
}
\institute{
KAIST \and Adobe Research}
\authorrunning{S. Woo et al.}
% \institute{
% KAIST \and Adobe Research}
% \end{comment}
%******************
\maketitle

\blfootnote{\textsuperscript{*} Work done during an internship at Adobe Research.}

\begin{abstract}
Scaling object taxonomies is one of the important steps toward a robust real-world deployment of recognition systems.
We have faced remarkable progress in images since the introduction of the LVIS benchmark.
To continue this success in videos, a new video benchmark, TAO, was recently presented.
Given the recent encouraging results from both detection and tracking communities, we are interested in marrying those two advances and building a strong large vocabulary video tracker.
However, supervisions in LVIS and TAO are inherently sparse or even missing, posing two new challenges for training the large vocabulary trackers.
First, no tracking supervisions are in LVIS, which leads to inconsistent learning of detection (with LVIS and TAO) and tracking (only with TAO).
Second, the detection supervisions in TAO are partial, which results in catastrophic forgetting of absent LVIS categories during video fine-tuning.
To resolve these challenges, we present a simple but effective learning framework that takes full advantage of all available training data to learn detection and tracking while not losing any LVIS categories to recognize.
With this new learning scheme, we show that consistent improvements of various large vocabulary trackers are capable, setting strong baseline results on the challenging TAO benchmarks.
\end{abstract}

\section{Introduction}

A central goal of computer vision is to produce a general-purpose perception system that robustly works in the wild.
Towards this ambitious goal, extending the current short category regime is one of the essential key milestones.
As an initial effort in this direction, the large-scale image benchmark, LVIS~\cite{gupta2019lvis}, was introduced and fostered significant progress in developing solid image domain solutions~\cite{kang2019decoupling,tan2020equalization,ren2020balanced,wang2021seesaw,tan2021equalization,zhou2022detecting}.
Recently, a video benchmark, TAO~\cite{dave2020tao}, calls for a shift from image to video, opening the new task of detecting and tracking large vocabulary objects.

\begin{figure*}[t]
    \centering 
    \includegraphics[width=0.8\textwidth]{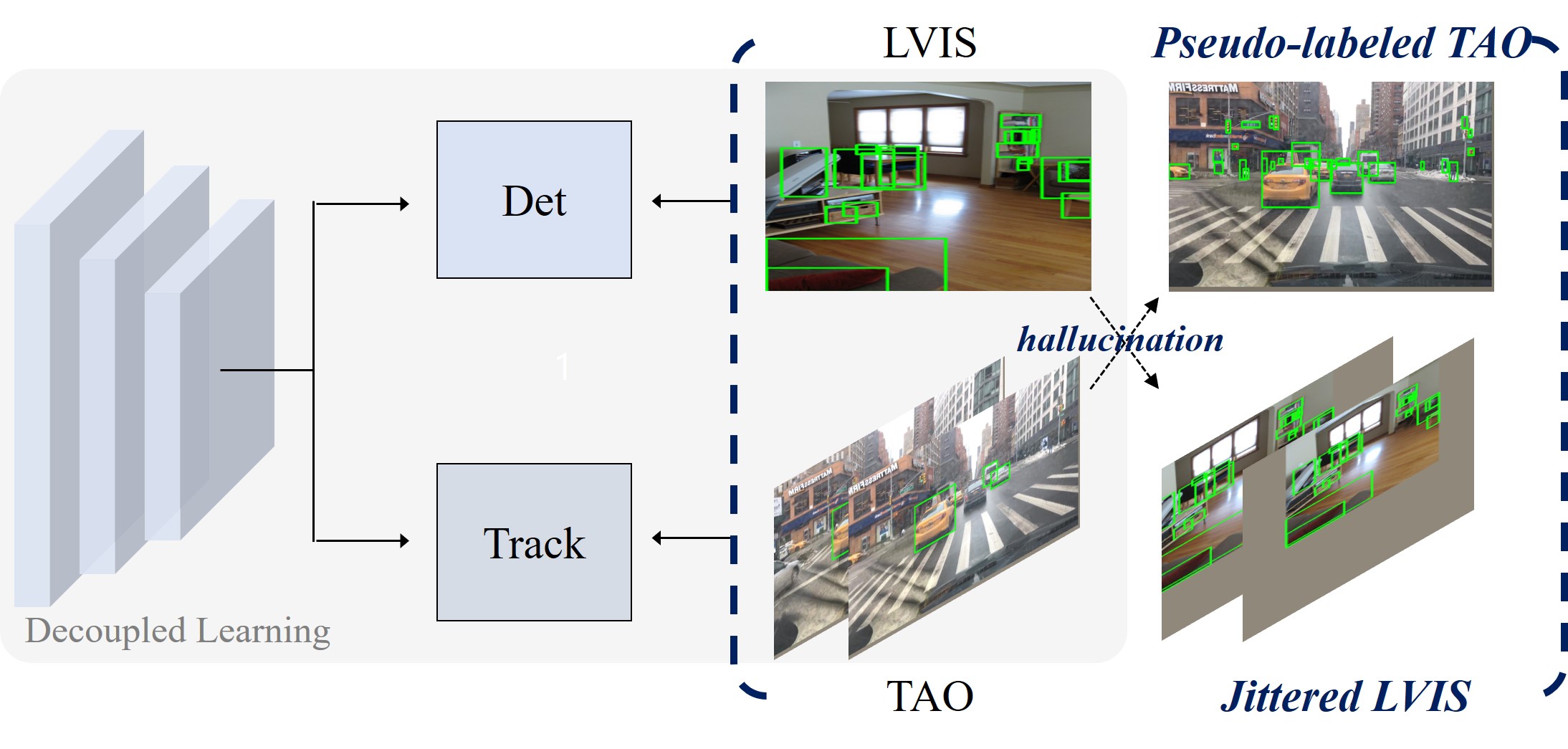}
    % \vspace{-3mm}
    \captionsetup{font=footnotesize}
    \caption{
    \textbf{The Proposed Learning Framework for Large Vocabulary Tracker Training.}
    While the current learning paradigm learns detection and tracking separately from LVIS and TAO (decoupled), our proposal takes all training data to learn detection and tracking jointly (\textbf{unified}). This is achieved through missing supervision hallucination.
    }
    % \vspace{-6mm}
    \label{fig:teaser}
\end{figure*}

With these new datasets of images and videos, LVIS and TAO, we are interested in building a strong large vocabulary video tracker. 
However, as the annotation difficulty between images and videos is even more severe in large vocabulary datasets, the significant gap in dataset scale and label vocabularies naturally exists.
Therefore, pre-training the model on images for learning large vocabularies and then fine-tuning on video for seamless video domain adaptation is a standard learning protocol.
Given this context, can the current advances of large-vocabulary detection and multi-object tracking be successfully unified and tied into a single model?
In particular, we see there are two main challenges for the successful marriage of two streams:
\textbf{First, no tracking supervisions are in LVIS.}
This essentially leads to inconsistent learning of detection (with LVIS and TAO) and tracking (only with TAO), resulting in sub-optimal video feature representations.
\textbf{Second, detection supervisions in TAO are partial}\footnote{
TAO dataset annotates 482 classes in total, which are the subset of LVIS dataset~\cite{gupta2019lvis}, and only 216 classes are in the training set.
}\textbf{.}
Thus, catastrophic forgetting~\cite{mccloskey1989catastrophic} is inevitable if one naively fine-tunes the LVIS tracker directly on the TAO.

In this work, we present simple, effective, and generic methods for hallucinating missing supervisions in each dataset. 
Below, we describe the challenges and our solutions in turn.

First, how can we simulate the tracking supervisions only with images in LVIS?
Given an image, our idea is to apply spatial jittering artifacts to mimic temporal changes in video and form a natural pair for tracking.
Here, we present two new spatial jittering methods. 
The first is strong zoom-in/out augmentation, which has a large scale-jittering effect that can effectively simulate the low sampling rate test-time inputs in large vocabulary tracking.
It yields significant performance improvements over the conventional image affine augmentation~\cite{held2016learning,oh2018fast,park2022per,zhou2020tracking,sio2020s2siamfc,zheng2021learning}.
Plus, our findings are in line with recent work that shows a large scale jittering is effective in image detection and segmentation~\cite{ghiasi2021simple}, and here we examine this observation on video for robust large vocabulary tracker training.
The second is mosaicing augmentation~\cite{yun2019cutmix,bochkovskiy2020yolov4}, which is originally presented for object detection with enriched background.
We extend this augmentation for combining foreground objects in different images in a class-balanced manner~\cite{gupta2019lvis} and simulate test-time hard, dense tracking scenarios suitable for ``many object" trackers.
We show that both are effective and complementary to each other.

Second, how can we fill the missing detection supervisions in TAO?
The TAO training data partially spans the LVIS categories, and thus the direct fine-tuning of LVIS tracker on TAO causes catastrophic forgetting of absent categories.
A straightforward way to avoid this issue is to learn only the tracking part of the model with TAO~\cite{pang2021quasi}.
However, this hinders full model training and abandons all the TAO detection labels, limiting the overall performance.
We instead approach this problem by combining the self-training~\cite{scudder1965probability,riloff1996automatically,riloff2003learning} with a teacher-student framework~\cite{hinton2015distilling,xie2020self}.
In practice, the teacher and student are identical copies of the LVIS pre-trained model, and we freeze the weights of the teacher during training.
The overall learning pipeline consists of two steps:
First, given an input, we predict pseudo labels using the teacher model.
The idea behind the pseudo labeling is to leverage the past knowledge acquired from LVIS and fill in the missing annotations in TAO.
Second, using the augmented labels, we train the student model with both distillation loss and ordinal detection loss.
Unlike the typical teacher-student schemes used in semi-supervised object detection studies~\cite{xu2021bootstrap,sohn2020simple,xu2021end}, we introduce two new adaptations suitable for large vocabulary learning setup.
The first is using \textit{soft} pseudo labels, i.e., distilling class logits directly, to fire all the student's classifier weights, rather than using common one-hot (hard) pseudo labels. This is crucial as standard hard pseudo labels tend to bias distillation only toward the frequent class objects due to inherent classifier calibration issue~\cite{dave2021evaluating,pan2021model}.
The second is to use \textit{MSE} loss in order to equally impact all the classifier weights~\cite{kim2021comparing,tang2021humble}, rather than using picky KL-divergence loss~\cite{hinton2015distilling}.
We found the type of the loss function is also very important for the successful large vocabulary classifier distillation.
Despite the simplicity, we empirically show that the distillation results are greatly improved with these adaptations.
%The student imitates the teacher's output (via distillation loss) to keep the ability to identify LVIS categories and learns from additional TAO detection labels (through ordinal detection loss).
We also show that our proposal works well on the common vocabulary setup, e.g., COCO, and can be easily extended to new class learning scenarios, COCO $\rightarrow$ YTVIS.
%successfully avoids catastrophic forgetting and benefits from additional labeled learning in TAO.
%, outperforming the previous naive approach significantly~\cite{pang2021quasi}.

Combining all these proposals together, unified learning of detection and tracking with both LVIS images and TAO videos becomes possible without forgetting any LVIS categories (see~\figref{fig:teaser}).
Furthermore, we also introduce a new regularization objective, semantic consistency loss.
It aims to prevent the common tracking failure in large vocabulary tracking due to semantic flicker between similar classes.
We study the efficacy of our final framework on the TAO benchmark and achieve new state-of-the-art results. 
Our extensive ablation studies confirm that the proposals are generic and effective.

\section{Related work}

\noindent{\textbf{Large vocabulary recognition.}} 
The object categories in natural images follow the Zipfian distribution~\cite{manning1999foundations}, and thus, the large vocabulary recognition is naturally tied with the long-tailed recognition~\cite{he2009learning,liu2019large,wu2020distribution}.  
Based on this connection, lots of solid approaches are introduced. The existing methods can be roughly categorized into \textit{data re-sampling} or \textit{loss re-weighting}.
The data re-sampling methods more often sample data from rare classes to balance the long-tailed training distribution~\cite{gupta2019lvis,chang2021image}. 
The loss re-weighting aims at adjusting the loss of each data instance based on their labels or train-time accumulated statistics~\cite{tan2020equalization,wang2021seesaw,tan2021equalization,hsieh2021droploss}.
Some approaches perform multi-staged training upon these methods, which first pre-train the model in a standard way and then fine-tune using either data re-sampling or loss re-weighting~\cite{kang2019decoupling,hu2020learning,wang2020frustratingly,li2020overcoming,ren2020balanced,wang2020devil,wang2021adaptive,zhang2021distribution}. Also, there are new approaches based on data augmentations~\cite{ghiasi2021simple,zang2021fasa,zhang2021mosaicos} or test time calibration~\cite{pan2021model}.

Apart from all these previous efforts on the image, we study a new video extension of the task~\cite{dave2020tao}. 
We show that our proposal is generic and not sensitive to a specific method, data re-sampling or loss re-weighting, in successfully converting the current large vocabulary detectors to large vocabulary trackers.

\vspace{2mm}
\noindent{\textbf{Multi-object tracking.}}
\sloppy Most modern multi object trackers~\cite{leal2017tracking} follow the tracking-by-detection paradigm~\cite{ramanan2003finding}.
An off-the-shelf object detector is first employed to localize all objects in each frame, and then track association is performed between adjacent frames.
The main difference among existing methods is in how they estimate the similarity between detected objects and previous tracks for the association.
To name a few, Kalman Filter~\cite{bewley2016simple,wojke2017simple}, optical flow~\cite{xiao2018simple}, displacement regression~\cite{zhou2020tracking,peng2020chained}, and appearance similarities~\cite{kim2015multiple,leal2016learning,leal2017tracking,milan2017online,son2017multi,sadeghian2017tracking,bergmann2019tracking,yang2019video,lu2020retinatrack,wang2020towards,zhang2021fairmot,pang2021quasi,zhang2021bytetrack} are the representatives.
On the other side, there are also efforts on joining detection and tracking~\cite{feichtenhofer2017detect,zhang2018integrated,wu2021track}, and recently by transformer-based architectures~\cite{sun2020transtrack,meinhardt2021trackformer,zeng2021motr}.
We note that all these methods only focus on a few object categories such as people or vehicles, ignoring the vast majority of objects in the world.

Our work is an early attempt for extending the current short category regime of modern trackers~\cite{dave2020tao,liu2021opening,wang2021unidentified,zhou2022global}.
In this paper, we build our proposal upon the tracking-by-detection paradigm.
We choose the state-of-the-art method, QDTrack~\cite{pang2021quasi}, which adopts Faster R-CNN~\cite{ren2015faster} and lightweight embedding head for detection and tracking, respectively.
The tracking is learned through a dense matching between quasi-dense samples on the pair of images and optimized with multiple positive contrastive learning.
Given the state-of-the-art large vocabulary detection~\cite{gupta2019lvis,tan2021equalization,wang2021seesaw} and multi-object tracking~\cite{pang2021quasi} methods, we primarily investigate the new challenges in developing a strong large vocabulary tracker.

%\jl{might give more explanation about QDTrack.}
\vspace{2mm}
\noindent{\textbf{Tracking without video annotations.}}
There is a line of recent research on self-supervised learning for tracking, either using unlabeled videos~\cite{vondrick2018tracking,wang2019learning,wang2019unsupervised,lai2019self,lai2020mast,li2019joint,purushwalkam2020aligning,xu2021rethinking} or images~\cite{held2016learning,oh2018fast,park2022per,zhou2020tracking,sio2020s2siamfc,fu2021learning,zheng2021learning}.
Our work belongs to the latter category.
%The idea is simple here.
Applying random affine augmentation to the original image provides a spatially jittered version, which mimics the temporal changes in the video.
By letting the model find the correspondence between those two images, meaningful tracking supervision can be provided~\cite{he2009learning,oh2018fast,zhou2020tracking}.
Sio~\etal\cite{sio2020s2siamfc} present that an image and any cropped region of it can generate a similar effect.
Zheng~\etal\cite{zheng2021learning} extend this idea to incorporate only the foreground objects in cropped regions for stable training.
%Recently, Fu~\etal\cite{fu2021learning} show that instance discrimination via contrastive loss within a single image can simulate tracking supervision.

In this work, we explore this general idea under a more specific large vocabulary tracking setting.
First, we focus on the fact that conventional motion cues are not applicable for large-vocabulray trackers as the input are temporally distant (1FPS) due to the annotation difficulties and there are severe camera movements in natural videos. This motivate us to train the tracker's vision feature matching more discriminative. To this end, we present a strong zoom-in/out augmentation that can not only simulate low sampling rate input but also includes large scale-jittering effect~\cite{ghiasi2021simple} which is known to be effective in the image domain vision tasks.
Second, we recast the image mosaicing augmentation~\cite{bochkovskiy2020yolov4}, which was initially proposed for robust object detection with enriched backgrounds~\cite{yun2019cutmix}, to simulate test-time dense tracking in the large vocabulary setting.
We show that both are complementary in providing discriminative tracking supervisions for this task.

\vspace{2mm}
\noindent{\textbf{Catastrophic forgetting.}}
The phenomenon wherein neural networks forget how to solve past tasks because of the exposure to new tasks is known as catastrophic forgetting~\cite{mccloskey1989catastrophic}. It occurs because the model weights that contain important information for the old task are over-written by information relevant to the new one.
While the catastrophic forgetting can occur in various scenarios, many existing efforts are focused on the class incremental learning setup, where it incrementally adds new object categories phase-by-phase, in image classification~\cite{kirkpatrick2017overcoming,lee2017overcoming,shin2017continual,rebuffi2017icarl,chaudhry2018riemannian,wu2019large,aljundi2019gradient,prabhu2020gdumb}.
Also, there are some few approaches tackling incremental object detection~\cite{kuznetsova2015expanding,shmelkov2017incremental,zhou2020lifelong,wang2021wanderlust}.

We target a different setup, transfer learning from image to video without forgetting.
Specifically, we aim to train the model on images covering the entire evaluation categories and then fine-tune it on videos, which partially covers the evaluation categories, without forgetting.
While lots of current video models~\cite{oh2019video,yang2019video,kim2020video} are trained in this way for generic feature learning, the label difference issue between images and videos has been rarely studied and explored.
We study this issue, as this is a practical setup for training large vocabulary trackers using both images and videos.

\begin{figure*}[t]
    \centering 
    \includegraphics[width=0.99\textwidth]{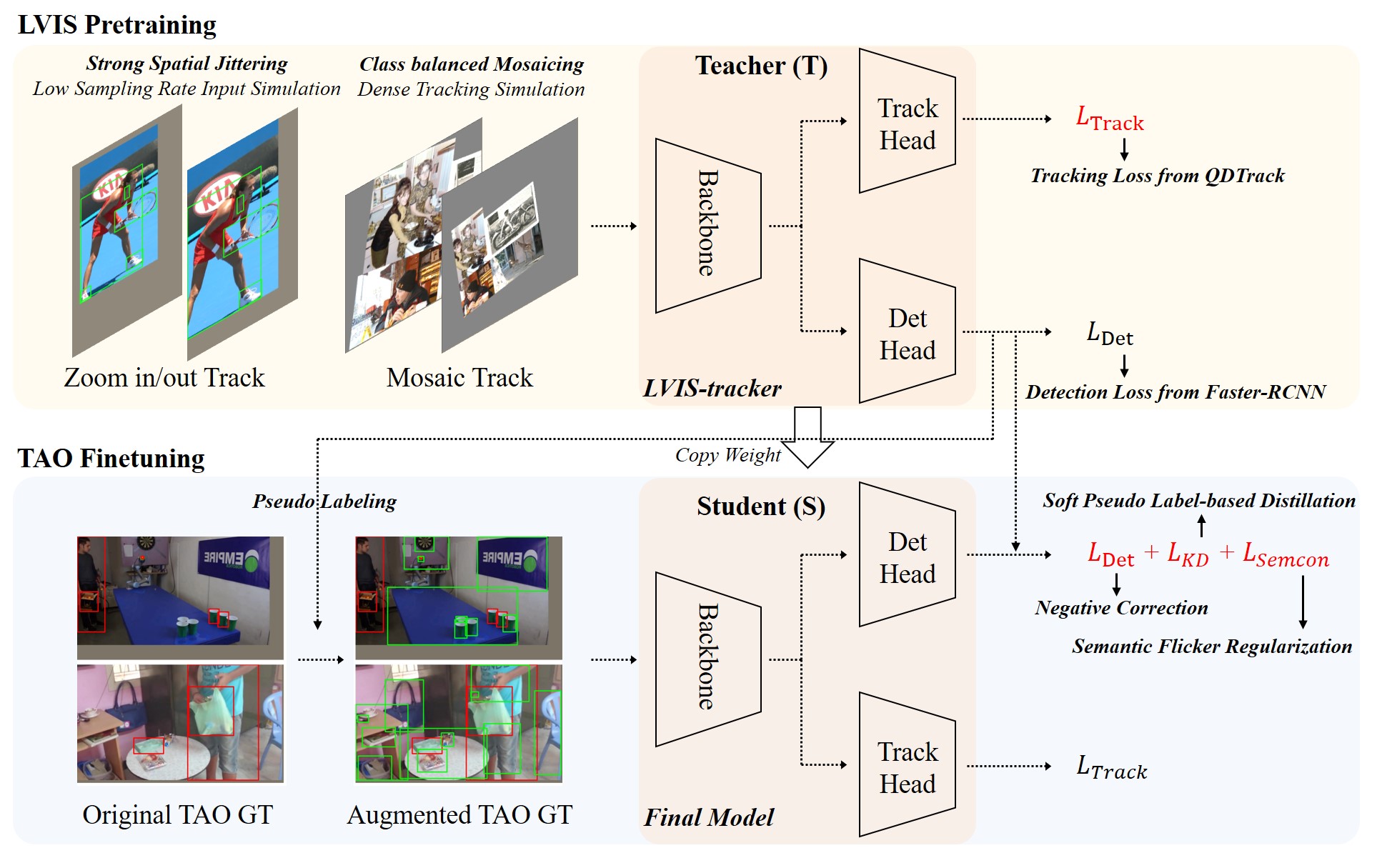}
    % \vspace{-2mm}
    \captionsetup{font=footnotesize}
    \caption{\textbf{Overview of the proposed learning framework.}
    The \textcolor{red}{red} colored objective functions are generated supervisions with our proposals.
    }
    \label{fig:main}
    % \vspace{-5mm}
\end{figure*}

\section{Proposed Method}

We introduce a general learning framework that allows joint learning of detection and tracking from all training data, LVIS and TAO, for robust large vocabulary tracking.
The overview of our pipeline is shown in~\figref{fig:main}.
We first present how we can learn tracking from images through zoom-in/out and mosaicing augmentations in~\secref{track_lvis}.
We then describe how we avoid catastrophic forgetting when the videos for fine-tuning have fewer label vocabularies than pre-trained images in~\secref{detect_tao}.
Finally, we present a new regularization loss term, namely semantic consistency loss, for preventing semantic flicker in \secref{sem_con}.

\subsection{Learn to Track in LVIS}
\label{track_lvis}

Our approach is straightforward.
An original image and a transformed image with the spatial jittering artifacts can form a natural input pair for tracking.
For the jittering artifacts, we present two new augmentations, zoom-in/out and mosaicing (see~\figref{fig:main}).
Note that tracking annotations come for free as we know the exact transformation relationship between the images.
We assign the same unique track-id to the same object in the transformed image.

\vspace{2mm}
\noindent\textbf{Strong Zoom-in/out Track.}
Due to the annotation difficulty of the large-vocabulary tracking dataset, the train and test time inputs are temporally sparse, i.e., low sampling rate, which naturally results in conventional motion cues not applicable and rather rely on pure vision feature matching.
To make the vision feature matching more discriminative, and to effectively simulate test-time low sampling rate inputs, we present strong zoom-in/out augmentation. 

It is mainly composed of scaling and cropping operations, which essentially vary the scale and position of the objects.
Specifically, for an image $\mathrm{I}$, we generate a input pair, $\mathrm{I_{t}}$ and $\mathrm{I_{t+\tau}}$, by applying the $\mathrm{scale\_and\_crop(\cdot)}$ function to each image.
In practice, it scales an image up to 2 times and crops the image to have a minimum IoU of 0.4 or above with original bounding boxes to avoid heavy object truncation and ensure stable tracker training.
Prior works either adopt standard random affine transformation~\cite{he2009learning,oh2018fast,zhou2020tracking} or cropping without scaling~\cite{sio2020s2siamfc,zheng2021learning}, which generally provide \textit{weak} scale-jittering effect.
Instead, we focus on enlarging the scaling effect and show that our proposal significantly outperforms the baselines.

\vspace{2mm}
\noindent\textbf{Mosaicing Track.}
%\subsubsection{Mosaicing augmentation}
While the zoom-in/out augmentation is already effective in providing tracking supervisions, it is limited in the tracking of a few objects due to the federated annotations of LVIS~\cite{gupta2019lvis}.
To resolve the issue, we present to combine multiple images and perform tracking with the increased foreground objects.
We implement our idea by extending the image mosaicing augmentation~\cite{bochkovskiy2020yolov4}, which stitches four random training images with certain ratios.
While it was originally presented for object detection with enriched background~\cite{yun2019cutmix}, we recast it to simulate hard, dense tracking scenarios in large vocabulary tracking.
In practice, four random images, $\{\mathrm{I_{a}, I_{b}, I_{c}, I_{d}}\}$, are sampled from RFS (Repeat Factor Sampling)-based dataset~\cite{gupta2019lvis} to maintain the class-balance.
Then, image stitching followed by random affine (with large scale jittering within a range of 0.1 to 2) and crop is applied.
We summarize these procedure as $\mathrm{mosaic(\cdot)}$. The tracking pair then can be obtained by applying the $\mathrm{mosaic(\cdot)}$ function to the sampled images twice.
However, we see that unnatural layout pair results in train and test time inconsistency.
To this end, we propose to sample tracking input pairs in a mixed way from two different augmentations, zoom-in/out and mosaicing, with equal probability during training. We empirically confirm that this works well in practice.

With our proposal, the model can receive tracking supervisions from all LVIS object categories. 
The tracking objective function is adopted from the QDtrack~\cite{pang2021quasi} (see~\figref{fig:main}-top), and we call this model LVIS-Tracker.
While the model is only trained on LVIS dataset, it already outperforms the previous state-of-the-art tracker (trained with the standard decoupled learning scheme) significantly (see~\tabref{tab:abl_sj}).

\subsection{Learn to Unforget in TAO}
\label{detect_tao}

Due to the fundamental annotation difficulties in videos, the images are in general bigger in dataset scale and larger in taxonomies.
Therefore, pre-training the model on images to acquire generic features and fine-tuning on videos for target domain adaptation has become a common protocol for obtaining satisfactory performance in various video tasks~\cite{oh2019video,yang2019video,kim2020video}.
This also applies to training the large vocabulary video trackers, where we first learn a large number of vocabulary from LVIS images and then adapt to the evaluation domain with TAO videos.
However, as TAO partially spans the full LVIS vocabularies, a naive transfer learning scheme results in catastrophic forgetting.
%Here, we present an effective solution to resolve this issue.

Here, our goal is to keep the ability to detect the previously seen object categories while also adapting to learn from new video labels.
We mainly focus on the catastrophic forgetting in the detector, as the tracking head is learned in a category-agnostic manner.
We detail the proposal using the standard two-staged Faster-RCNN detector (FPN backbone)~\cite{ren2015faster,lin2017feature}.
Without loss of generality, the proposals can be extended to multi-staged architectures~\cite{cai2018cascade,cai2019cascade,chen2019hybrid,vu2019cascade}, where we apply the proposal for each RCNN head and average them.
In fact, the main issue is missing annotations for the seen, known object categories during the image to video transfer learning.
Since they are not annotated, we can neither provide detection supervision nor prevent them from being treated as background.
This basically perturbs the pre-trained classifier boundaries of both RPN and RCNN, leading to catastrophic forgetting.
We remedy this issue by presenting a pseudo-label guided teacher-student framework.

Our key idea is intuitive.
The pre-trained model already has sufficient knowledge to detect the seen, known categories.
Based on this fact, we first fill in the missing annotations by pseudo-labeling the input.
We adopt the basic pseudo-labeling scheme with a threshold of 0.3.
The redundant pseudo labels that highly overlap with the current labels are filtered out with NMS.
With these augmented labels, we 1) design a teacher-student network to provide (soft) supervisions, i.e., class logit, for preserving the past knowledge, and
2) update the incorrect background samples, i.e., negatives, in RPN and RCNN to prevent seen objects from being background (see~\figref{fig:main}-bottom).
Using soft class logit is important for the large vocabulary classifier distillation, as the hard pseudo labels bias the operation towards the frequent class objects.
Moreover, we use MSE loss instead of Kullback-Leibler (KL) divergence loss~\cite{hinton2015distilling} for the logit matching. This is because the MSE loss treats all classes equally and thus it allows the rare classes with low probability also to be updated properly~\cite{tang2021humble}. This two new adaptation leads to the successful distillation of the previous knowledge of the large vocabulary classifier (see~\tabref{tab:abl_ts_abl}).

\vspace{2mm}
\noindent\textbf{Teacher-Student Framework Setup.}
To effectively retain the previous knowledge, we design a teacher-student framework.
We first make identical copies of the image pre-trained model, teacher (T) and student (S).
The teacher model (T) is frozen to keep the previous knowledge and guide the student.
The student model (S) adapts to the new domain with incoming video labels (via detection loss) and also mimics the teacher model to preserve the past information (via distillation loss).
We detail the components in the following.
%The distillation targets are sampled based on the augmented labels; 

\vspace{2mm}
\noindent\textbf{RPN Knowledge Distillation Loss.}
The RPN takes multi-level features from the ResNet feature pyramid~\cite{lin2017feature}.
In particular, each feature map is embedded through the convolution layer, followed by two separate layers, one for objectness classification and the other for proposal regression.
We collect the outputs of both heads from the teacher and student to compute RPN distillation loss, which is defined as
$
% \begin{equation}
    % \begin{split}
\mathit{L}^\mathrm{RPN}_\mathrm{KD} = \frac{1}{\mathit{N}_{cls}}{\sum_{i=1} \mathit{L}_{cls}(u_{i}, u_{i}^{*})} 
                                    + \frac{1}{\mathit{N}_{reg}}{\sum_{i=1} \mathit{L}_{reg}(v_{i}, v_{i}^{*})}.
    % \end{split}
$
% \end{equation}
%
Here, $i$ is the index of an anchor. 
$u_{i}$ and $u_{i}^{*}$ are the mean subtracted objectness logits obtained from the student and the teacher, respectively.
$v_{i}$ and $v_{i}^{*}$ are four parameterized coordinates for the anchor refinement obtained from the student and teacher, respectively.
$\mathit{L}_{cls}$ and $\mathit{L}_{reg}$ are MSE loss and smooth L1 loss, respectively.
Here, we note that $\mathit{L}_{reg}$ is only computed for the positive anchors that have an IoU larger than 0.7 \textit{with the augmented ground-truth boxes}.
$\mathit{N}_{cls} (=256)$ and $\mathit{N}_{reg}$ are the effective number of anchors for the normalization.

\vspace{2mm}
\noindent\textbf{RCNN Knowledge Distillation Loss.} 
We perform RoIAlign~\cite{he2017mask} on top-scoring proposals from RPN, extracting the region features from each feature pyramid level. 
Each region feature is embedded through two FC layers, one for classification and the other for bounding box regression. We collect the outputs of both heads from the teacher and student to compute RCNN distillation loss, which is defined as
$
% \begin{equation}
%     \begin{split}
    \mathit{L}^\mathrm{RCNN}_\mathrm{KD} = \frac{1}{\mathit{M}_{cls}}{\sum_{j=1} \mathit{L}_{cls}(p_{j}, p_{j}^{*})} 
                                            + \frac{1}{\mathit{M}_{reg}}{\sum_{j=1} \mathit{L}_{reg}(t_{j}, t_{j}^{*})}. 
%     \end{split}
% \end{equation}
%
$
Here, $j$ is the index of a proposal. 
$p_{j}$ and $p_{j}^{*}$ are the mean subtracted classification logits obtained from the student and the teacher, respectively.
$t_{j}$ and $t_{j}^{*}$ are four parameterized coordinates for the proposal refinement obtained from the student and teacher, respectively.
$\mathit{L}_{cls}$ and $\mathit{L}_{reg}$ are MSE loss and smooth L1 loss, respectively.
We only impose $\mathit{L}_{reg}$ for the positive proposals that have an IoU larger than 0.5 \textit{with the augmented ground-truth boxes}.
$\mathit{M}_{cls} (=512)$ and $\mathit{M}_{reg}$ are the effective number of proposals for the normalization.

\begin{figure}[t]
    \centering 
    \vstretch{0.95}{\includegraphics[width=0.95\textwidth]{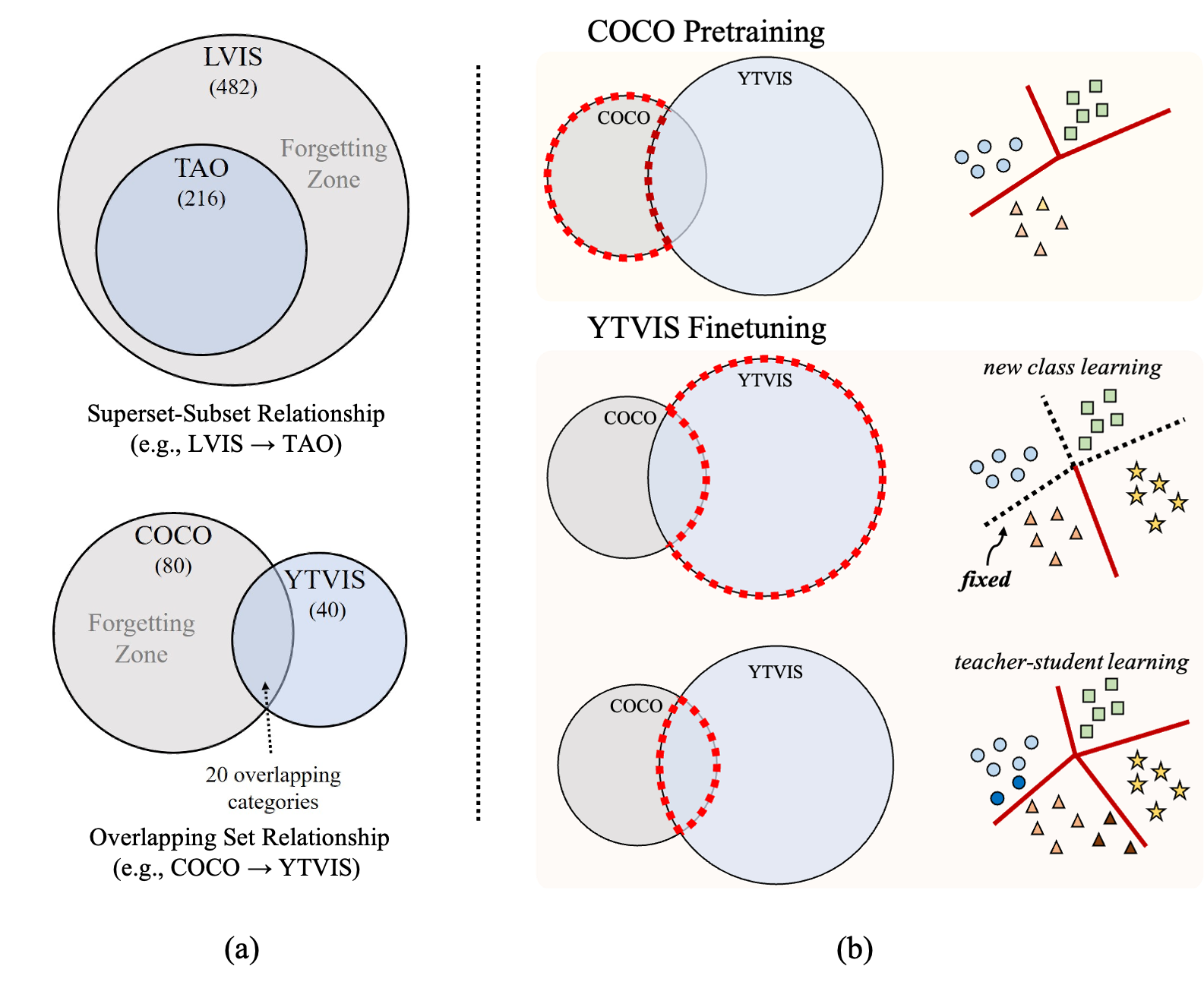}}
    \vspace{-2mm}
    \captionsetup{font=footnotesize}
    \caption{
    \textbf{(a) Two standard image to video transfer learning setups.}
    In typical, a naive transfer learning from images to videos leads to catastrophic forgetting due to the missing annotations in the video.
    We present a generic teacher-student scheme that works on both scenarios.
    \textbf{(b) Two-step approach for COCO $\rightarrow$ YTVIS transfer learning setup.}
    We first learn new object classifier weights with the pre-trained classifier as a fixed anchor and then fine-tune the whole classifier through the proposed teacher-student scheme.
    The red-dotted line along the circle in the set relationship figure indicates the training data used in each stage.
    The shape figures (e.g., square, triangle) and the separating line denote class instances and the associated classifier.
    }
    \vspace{-3mm}
    \label{fig:im_to_vid_rel}
\end{figure}

\vspace{2mm}
\noindent\textbf{Correcting Negatives in Computing the Detection Loss.}
We avoid sampling the anchors or proposals that have significant IoU overlaps \textit{with the augmented ground-truth boxes} as a background ($>$ 0.7 for RPN and $>$ 0.5 for RCNN).
We note that positives are only sampled based on the provided original ground truth labels.
This is because the detectors, especially the large vocabulary detectors, suffer from predicting the precise labels~\cite{dave2021evaluating,pan2021model} while they are good at recalling the objects. We empirically verify this in the experiment.

\vspace{2mm}
\noindent\textbf{Extension to other transfer learning setup.}
COCO to YTVIS is another important transfer learning setup (see~\figref{fig:im_to_vid_rel}-(a)).
This is more challenging than LVIS to TAO, as the superset-subset relationship does not hold, and new object categories to learn are added.
To deal with this new pattern, we take a two-step approach (see~\figref{fig:im_to_vid_rel}-(b)).
First, we adapt the RCNN classifier of the pre-trained model, increasing the number of output channels to accommodate newly added classes, and train on the videos, $\mathrm{YTVIS} - \mathrm{COCO}$, that contain new object categories. 
In practice, we freeze the original detector, and thus the past information is intact, and only the newly added weight matrices are updated accordingly.
The key idea here is to use the original pretrained weight as an anchor and update the newly added weight to be compatible.
Second, after sufficient training of the new weights, we now unlock the original detector and update the whole weights with the remaining videos, $\mathrm{YTVIS} \cap \mathrm{COCO}$, using the presented teacher-student scheme.

\subsection{Regularizing Semantic Flickering}
\label{sem_con}
One of the common tracking failures in large vocabulary tracking is due to semantic flicker between similar object categories~\cite{dave2020tao}.
To cope with this issue, we attempt to regularize the model during training with a new objective function, namely semantic consistency loss.
The proposal is motivated by the temporal consistency loss~\cite{barnes2009patchmatch,lai2018learning,kim2019deep,lei2020blind}, which enforces the outputs of the model for corresponding pixels (or patches) in video frames to be consistent. It is often used in video processing tasks to ensure the output temporal smoothness at a pixel level.
The proposal extends this idea from pixels to instances; We enforce the class predictions of the same instances in two different frames to be equivalent.
In practice, we forward the ground truth bounding boxes of the same instance in two different frames to the RCNN head.
The mean subtracted classification logits, $p$, are used for the consistency regularization as,
$
% \begin{equation}
% \begin{split}
    \mathit{L}_\mathrm{Semcon} = |p^{t} - p^{t+\tau}|_{2}.
% \end{split}
% \end{equation}
$
ere, $p^{t}$ and $p^{t+\tau}$ denote the logits of the same instance in two different frames, $I_{t}$ and $I_{t+\tau}$.

\subsection{Unified Learning}
Within our proposed learning framework (see~\figref{fig:main}), we can train the whole video model, learning detection and tracking jointly, using all available image and video datasets.
The final objective function can be summarized as
\begin{equation}
\begin{split}
    \mathit{L} =  \lambda_\mathrm{1} \mathit{L}_\mathrm{Det} + \lambda_\mathrm{2} \mathit{L}_\mathrm{Track} + \lambda_\mathrm{3} \mathit{L}_\mathrm{KD} + \lambda_\mathrm{4} \mathit{L}_\mathrm{Semcon},
\end{split}
\end{equation}
which consists of four loss terms in total.
The detection ($\mathit{L}_\mathrm{Det}$) and tracking losses ($\mathit{L}_\mathrm{Track}$) are adopted from \cite{ren2015faster} and \cite{pang2021quasi}.
Note that the $\mathit{L}_\mathrm{KD} (=\mathit{L}^\mathrm{RPN}_\mathrm{KD} + \mathit{L}^\mathrm{RCNN}_\mathrm{KD}) $ and $\mathit{L}_\mathrm{Semcon}$ are used only when fine-tuning on the videos.

\section{Experiments}

In this section, we conduct extensive experiments to analyze our methods.
We investigate the results mainly in two aspects: image-level prediction and cross-frame association, which will be reflected in the BBox AP and Track AP~\cite{dave2020tao}, respectively. Considering the task difficulty, we mainly focus on the Track AP of 50, 75 and their average.
For the TAO test, we provide Track AP*, a full Track AP average for IoU from 0.5 to 0.95 with a step size of 0.05.
We study the impact of unified learning on TAO dataset (\secref{sec:main_results}). We consistently outperformed the current decoupled learning paradigm with healthy margins using various models, and pushed the state-of-the-art performance significantly.
Second, to investigate the importance of the major components in our proposals, we provide ablation studies on TAO validation set (\secref{sec:ablation}).
Lastly, we evaluate our teacher-student scheme on two representative image-video transfer learning scenarios, LVIS $\rightarrow$ TAO and COCO $\rightarrow$ YTVIS\footnote{For the experiment, we contact the authors for the YTVIS-\textit{val} annotations.}(\secref{sec:transfer}).
In the following, we provide experiment setups, evaluation protocol and results for each section. More details are in supplementary materials.

\subsection{Main Results}
\label{sec:main_results}

Upon the state-of-the-art tracking-by-detection framework~\cite{pang2021quasi}, we instantiate various large vocabulary trackers.
In specific, we consider two important detection architecture, two-staged (Faster-RCNN~\cite{ren2015faster}) and multi-staged (CenterNet2~\cite{zhou2021probabilistic}), and three different long-tailed learning methods, Repeat Factor Sampling (RFS)~\cite{gupta2019lvis}, Equalization Loss V2 (EQLv2)~\cite{tan2021equalization}, and Seesaw Loss~\cite{wang2021seesaw}.
All the models use the same ResNet-101~\cite{he2016deep} with feature pyramid~\cite{lin2017feature} backbone following the previous works~\cite{dave2020tao,pang2021quasi}.
Based on these baseline models, we compare our learning framework with the current standard learning protocol, decoupled learning.
The comparison is in~\tabref{tab:tao_main}.
We observe that our unified learning scheme consistently outperforms the current decoupled learning paradigm on various models, showing the strong generalizabilty of the proposal.
With our method, we push the state-of-the-art performance significantly, achieving 21.6 and 20.1 Track AP50 on TAO-\textit{val} and TAO-\textit{test}, respectively.

\begin{table*}[t]
 \centering
  \subfloat[\scriptsize \textbf{State-of-the-art results} in TAO-\textit{val}.]{
    \label{tab:tao_val}
    \resizebox{0.47\textwidth}{!}
    {
    \def\arraystretch{0.8}
    \begin{tabular}{lccr}
    \toprule
    Method                               & Track $\mathrm{AP}_{50}$ & Track $\mathrm{AP}_{75}$ & Track $\mathrm{AP}_{avg}$                                \\
    \midrule
    SORT\_TAO~\cite{dave2020tao}         & 13.2 & - & -                                \\
    FasterRCNN-RFS~\cite{pang2021quasi}  & 16.1 & 5.0 & 10.6                            \\
    \midrule
    FasterRCNN-RFS*                      & 13.4 & 4.9 & 9.2                            \\
    w/ SimLearn                          & \textbf{19.6} &\textbf{ 7.3} & \textcolor{blue}{($+$4.4)} \textbf{13.6}  \\
    \midrule
    FasterRCNN-EQLv2                     & 14.2 & 5.5 & 10.1                           \\
    w/ SimLearn                          & \textbf{19.8} & \textbf{8.8} & \textcolor{blue}{($+$4.1)} \textbf{14.2}  \\
    \midrule
    FasterRCNN-Seesaw                    & 15.4 & 5.7 & 10.5                           \\
    w/ SimLearn                          & \textbf{20.2} & \textbf{9.4} & \textcolor{blue}{($+$4.3)} \textbf{14.8}  \\
    \midrule
    CenterNet2-RFS                       & 18.9 & 9.1 & 14.0                           \\
    w/ SimLearn                          & \textbf{21.6} & \textbf{10.4} & \textcolor{blue}{($+$2.1)} \textbf{16.1} \\
    \bottomrule
    \end{tabular}
    }
 }
  \subfloat[\scriptsize \textbf{State-of-the-art results} in TAO-\textit{test}.]{%
    \label{tab:tao_test}
    \resizebox{0.51\textwidth}{!}
    {
    \def\arraystretch{1.75}
    \begin{tabular}{lccr}
    \toprule
    Method                        & Track $\mathrm{AP}_{50}$ & Track $\mathrm{AP}_{75}$ & Track $\mathrm{AP}_{50:95}$ \\
    \midrule
    SORT\_TAO~\cite{dave2020tao}  &  10.2      & 4.4        & 4.9      \\
    FasterRCNN-RFS~\cite{pang2021quasi}  &  12.4      & 4.5        & 5.2      \\
    \midrule
    FasterRCNN-RFS* + SimLearn.        & 17.1      & 6.9         & 7.8      \\
    FasterRCNN-EQLv2 + SimLearn.       & 16.8      & 7.2         & 8.0      \\
    FasterRCNN-Seesaw + SimLearn.      & 17.6      & 8.0         & 8.5      \\
    CenterNet2-RFS + SimLearn.         & 20.1      & 9.6         & \textbf{10.3}     \\
    \bottomrule
    \end{tabular}
    }
 }
\vspace{2mm}
\captionsetup{font=footnotesize}
\caption{Our learning framework couples well with different model architectures and learning methods. 
All the baseline scores are obtained after the decoupled training, i.e., training the detector and tracker on LVIS and TAO, respectively.
FasterRCNN-RFS* is a re-implementation of~\cite{pang2021quasi} baseline.
}
\label{tab:tao_main}
\vspace{-6mm}
\end{table*}

\subsection{Ablation studies}
\label{sec:ablation}

\vspace{2mm}
\noindent \textbf{Impact of image spatial jitterings.}
The results are presented in~\tabref{tab:abl_sj}.
Compared to the standard affine transformation~\cite{he2009learning,oh2018fast,zhou2020tracking} or simple cropping without scaling~\cite{sio2020s2siamfc,zheng2021learning}, the presented strong zoom-in/out and mosaicing provides a large Track AP improvement.
This indicates both the low sampling rate input simulation (with large-scale jittering) and dense tracking simulation (with mosaicing) enables more accurate large vocabulary object associations at test-time.
To concretely investigate the scaling effects of zoom-in/out, we also provide its variant with small scale-jittering, Z-in/out*, and confirm that the large scale-jittering~\cite{ghiasi2021simple} is indeed important for the performance.
We notice that mosaicing augmentation drops the Box AP.
We conjecture this happens due to the train and test time inconsistency of input pairs.
To this end, we present to form a tracking pair from two different augmentations in equal probability.
We found that this mixed sampling strategy provides the best Track AP.

\begin{table*}[t]
 \centering
  \subfloat[\centering \scriptsize \textbf{Effect of Spatial Jittering Strategies} in LVIS pre-train.]{%
    \label{tab:abl_sj}
    \resizebox{0.49\textwidth}{!}
    {
    \def\arraystretch{1.15}
    \begin{tabular}{l|ccc}
    \toprule
    Method                                                                  & Box AP    & Track $\mathrm{AP}_{50}$ & Track $\mathrm{AP}_{75}$ \\
    \midrule
    Decoupled TAO-tracker~\cite{pang2021quasi}                              & 18.1      & 16.1       & 5.0      \\
    \midrule
    Random Affine~\cite{held2016learning,oh2018fast,zhou2020tracking}       & 17.4      & 13.6       & 5.0      \\
    Random Crop~\cite{sio2020s2siamfc,zheng2021learning}                    & 15.9      & 11.5       & 1.5      \\
    Zoom in/out*                                                            & 18.1      & 13.2       & 4.4      \\
    \midrule
    Zoom in/out                                                             & \textbf{19.0} & 14.4          & \textbf{6.0}     \\
    Mosaic                                                                  & 16.2          & 16.5          & 4.7     \\
    Both (LVIS-tracker)                                                     & 18.5          & \textbf{17.8} & 5.7     \\
    \bottomrule
    \end{tabular}
    }
}
  \subfloat[\centering \scriptsize \textbf{Effect of Teacher-Student Framework} in TAO fine-tune.]{%
    \label{tab:abl_ts_abl}
    \resizebox{0.49\textwidth}{!}
    {
    \def\arraystretch{0.9}
    \begin{tabular}{l|ccc}
    \toprule
    Method                          & Box AP & Track $\mathrm{AP}_{50}$ & Track $\mathrm{AP}_{75}$      \\
    \midrule
    LVIS-tracker                    & 18.5   & 17.8       & 5.7          \\
    \midrule
    Naive-ft                        & 11.7   & 11.4       & 2.7           \\
    \midrule
    Vanilla Teacher-Student         & 18.0   & 15.9       & 6.7          \\
    + Pseudo-distill target         & 18.6   & 17.8       & 7.2          \\
    + Pseudo-neg sample             & 18.8   & 17.9       & 7.3          \\
    Both (Ours)                     & 19.5   & 18.5       & \textbf{7.5}          \\
    \midrule
    w. Hard Pseudo label~\cite{xu2021bootstrap,sohn2020simple,xu2021end}       & 17.1   & 15.9       & 5.3        \\
    w. KL-based distill~\cite{hinton2015distilling}                             & 17.7   & 16.8       & 6.0          \\
    \midrule
    w. Semcon. (Final Tracker)                 & \textbf{19.6}   & \textbf{19.5}       & 7.3 \\
    \bottomrule
    \end{tabular}
    }
}
\vspace{2mm}
\captionsetup{font=footnotesize}
\caption{(a) Zoom-in/out* and Zoom-in/out denote zoom-in/out augmentation with scaling range of [0.8, 1.25] and [0.1, 2.0], respectively.
(b)Pseudo Labeled Training denotes the standard (hard) pseudo label-based training.}
\label{tab:tao_abl}
\vspace{-3mm}
\end{table*}

\vspace{2mm}
\noindent \textbf{Impact of teacher-student framework.}
In~\tabref{tab:abl_ts_abl}, we study the impact of the key proposals in teacher-student framework.
For the baselines, we provide the Naive-ft and Vanilla Teacher-Student schemes.
Naive-ft indicates fine-tuning on TAO videos without any proper regulation for forgetting, which results in a significant performance drop.
Vanilla Teacher-Student scheme samples the distillation targets only from the original ground truth labels, and no negative correction is performed. While it shows the past knowledge preservation effect to some extent, the performance is still worse than the LVIS-tracker.
The vanilla scheme starts to improve over the LVIS-tracker when our proposal is added.
This implies that pseudo labeling is essential, and 1) keeping the past knowledge of seen objects (by sampling distillation targets from the augmented labels) and 2) preventing the seen objects from being background (by correcting negatives using the augmented labels) are the key to avoid catastrophic forgetting.

One may wonder if the standard (hard) pseudo-labeling approach can directly preserve the previous knowledge as typical teacher-student scheme do~\cite{xu2021bootstrap,sohn2020simple,xu2021end}.
However, as can be shown in the results, we instead observe inferior results than the baseline.
The large vocabulary classifier fundamentally suffers from the confidence calibration issue~\cite{dave2021evaluating,pan2021model} as it is trained on the long-tailed class-imbalanced data. 
It results in the classifier bias; predictions are made mainly toward the frequent object categories, missing rare objects in one-hot hard pseudo labels. 
In contrast, the (soft) pseudo labels essentially affect all classes.
Furthermore, we suggest to employ MSE loss rather than standard KL-loss~\cite{hinton2015distilling} as objective function in distillation. 
As MSE loss treats all classes equally the impact of the gradient is not attenuated for the rare classes.
Recent study also reveals that MSE loss offers better generalization capability due to the direct matching of logits compared to the KL loss~\cite{kim2021comparing}.

\vspace{2mm}
\noindent \textbf{Impact of semantic consistency loss.}
Finally, we study the impact of semantic consistency loss.
It regularizes the model's class logits of the same instance in different frames to be the equivalent.
In~\tabref{tab:abl_ts_abl}, we observed meaningful improvement in Track AP.
This implies that semantic flicker regularization is indeed effective for the large vocabulary object tracking.

\subsection{Image to Video Transfer Learning}
\label{sec:transfer}
Here, we evaluate our teacher-student scheme on two representative image to video transfer learning setups (see~\figref{fig:im_to_vid_rel}).
In LVIS $\rightarrow$ TAO setup, we pre-train FasterRCNN-RFS tracker on LVIS (with 482 categories) and fine-tune on TAO (with 216 categories).
We evaluate the model on TAO-\textit{val} with Track AP metric.
In COCO $\rightarrow$ YTVIS setup, we pre-train Mask-RCNN~\cite{he2017mask} on COCO, transfer the weights to MaskTrack RCNN~\cite{yang2019video}, add new randomly initialized classifier weights to accommodate newly added classes, and fine-tune on YTVIS. 
More details of the setup are in supplementary materials.
We evaluate the model on YTVIS-\textit{val} with Mask AP~\cite{yang2019video} metric.
To quantitatively analyze whether the proposal properly preserves the past knowledge and benefits from the new video labels, we provide the scores of OLD and NEW.
Here, OLD indicates the classes that only reveal in the image pre-training stage.
NEW denotes the classes that appears in the video fine-tuning stage.
For each setup, we provide a baseline of naive fine-tuning, which results in a severe catastrophic forgetting.
The results are summarized in~\tabref{tab:lvis_to_tao} and \tabref{tab:coco_to_ytvis}.

\begin{table*}[t]
\centering
\resizebox{0.95\textwidth}{!}
{
\def\arraystretch{1.0}
\begin{tabular}{l|ccc | ccc | ccc }
\toprule
\multirow{2}{*}{Method} & \multicolumn{3}{c}{OLD (LVIS $-$ TAO)} & \multicolumn{3}{c}{NEW (TAO)} & \multicolumn{3}{c}{ALL}\\ 
                        \cline{2-4} \cline{5-7} \cline{8-10} 
                        & Box AP & $\mathrm{Track AP}_{50}$ & $\mathrm{Track AP}_{75}$ & Box AP & $\mathrm{Track AP}_{50}$ & $\mathrm{Track AP}_{75}$ & Box AP & $\mathrm{Track AP}_{50}$ & $\mathrm{Track AP}_{75}$  \\ 
\midrule
LVIS-tracker            & 15.7   & 16.1          & 5.7        & 21.1   & 17.2         & 9.0       & 18.5.  & 17.8          & 5.8 \\
\midrule
Naive-ft                & 7.1    & 7.7           & 1.3        & 16.2   & 14.9         & 3.7       & 11.7.  & 11.4          & 2.6 \\
Track-only              & 15.7   & 15.3          & 5.5        & 21.1   & 16.9         & 7.1       & 18.5.  & 16.1          & 6.3 \\
Teacher-Student         & \textbf{15.7}   & \textbf{16.3}  & \textbf6.5 & \textbf{23.1}   & \textbf{20.6}   & \textbf{9.0}  & \textbf{19.5} & \textbf{18.5} & \textbf{7.5} \\ 
\bottomrule
\end{tabular}
}
\vspace{2mm}
\captionsetup{font=footnotesize}
\caption{
Teacher-Student framework in \textbf{LVIS $\rightarrow$ TAO} transfer learning setup.
Evaluated on TAO-\textit{val}.
}
\vspace{-3mm}
\label{tab:lvis_to_tao}
\end{table*}

\begin{table*}[t]
\centering
\resizebox{0.96\textwidth}{!}
{
\def\arraystretch{1.1}
\setlength{\tabcolsep}{10pt}
\begin{tabular}{l|ccc | ccc | ccc }
\toprule
\multirow{2}{*}{Method} & \multicolumn{3}{c}{OLD (COCO $-$ YTVIS)} & \multicolumn{3}{c}{NEW (YTVIS)} & \multicolumn{3}{c}{ALL}\\ 
                        \cline{2-4} \cline{5-7} \cline{8-10} 
                        & AP$_{50}$ & AP$_{75}$ & AP & AP$_{50}$ & AP$_{75}$ & AP & AP$_{50}$ & AP$_{75}$ & AP  \\ 
\midrule
Naive-ft                 & 5.2 & 2.6 & 2.7 & 38.2 & 21.0 & 20.6 & 30.0 & 16.4 & 16.1 \\
Teacher-Student (1-step) & 38.9 &31.2 &26.2 &36.2 &19.4 &20.1 &36.9 &22.4 &21.6 \\
Teacher-Student (2-step) & 51.0 & 44.3 & 36.1 & \textbf{43.3} & 23.8 & \textbf{23.8} & \textbf{45.2} & 28.9 & 26.9 \\
\midrule
Full-ft (Oracle)        & \textbf{54.9} & \textbf{46.9} & \textbf{38.9} & 40.7 & \textbf{25.1} & 23.3 & 44.3 & \textbf{30.6} & \textbf{27.2} \\
\bottomrule
\end{tabular}
}
\vspace{2mm}
\captionsetup{font=footnotesize}
\caption{
Teacher-Student framework in \textbf{COCO $\rightarrow$ YTVIS} transfer learning setup.
Evaluated on YTVIS-\textit{val}.
}
\vspace{-6mm}
\label{tab:coco_to_ytvis}
\end{table*}

\vspace{2mm}
\noindent\textbf{LVIS $\rightarrow$ TAO Transfer Learning.}
Especially in this setup, all necessary vocabularies are already learned at the image pre-training stage. Therefore, we can avoid catastrophic forgetting by fine-tuning only the tracking part.
However, as it only updates the video model partially, it leads to inconsistent video representations, and thus performance rather slightly drops from the baseline LVIS-tracker.
Instead, our method preserves the performance of OLD classes (preventing catastrophic forgetting) and significantly improves the NEW class performance (benefiting from labeled learning).

\vspace{2mm}
\noindent\textbf{COCO $\rightarrow$ YTVIS Transfer Learning.}
This setup is more challenging as the model is required to achieve two goals, new class learning and old class preserving simultaneously.
We decompose these goals and approach this setup in two-step as described in~\secref{detect_tao}.
As can be shown in the results, the proposed two-step approach performs better than the direct application of teacher student scheme.
The final performance is comparable with, and in NEW classes outperforms, the oracle setup that use all the YTVIS training videos.
This shows that our proposal is generic and effective for standard image to video transfer learning setups.

\section{Conclusion}
In this paper, we tackle the challenging problem of learning a large vocabulary video tracker.
We present a simple learning framework that uses all LVIS images and TAO videos to jointly learn the detection and tracking.
In specific, first, two spatial jittering methods, strong zoom-in/out and mosaicing, which effectively simulate the test-time large vocabulary object tracking are presented to enable tracker training with LVIS.
Second, a generic teacher-student scheme is proposed to prevent catastrophic forgetting while fine-tuning the image pre-trained models on videos.
We show that two new adaptation of using soft labels with MSE loss is crucial for the large vocabulary classifier distillation.
We hope our new learning framework settles as a baseline learning scheme for many follow-up large-vocabulary trackers in the future.

\vspace{3mm}
\noindent\textbf{Acknowledgement}
This work was supported in part supported by Samsung Electronics Co., Ltd (G01200447)

\appendix
\section{Appendix}

\noindent In this appendix, we provide,
\begin{enumerate}[label=A\arabic*.]
    \item Our view of the proposal from video data scaling perspectives,
    \item Datasets specifics used in the experiments,
    \item Implementation details including COCO $\rightarrow$ YTVIS transfer learning setup,
    \item Oracle analyses to investigate the disentangled impact of the method on object classification and tracking,
\end{enumerate}

\subsection{Bridging Images and Videos}
Applying deep learning in the video domain fundamentally suffers from the data-hungry issue, and the situation will become even more severe for more complex and challenging tasks.
One promising direction we believe is leveraging already well-curated large-scale image data to complement the insufficient video data.
However, jointly using multiple datasets~\cite{zhou2022simple}, image and video labels, leads to several issues, detailed below.

In this paper, we investigate the new problem of large vocabulary tracking, one of the essential milestones for dynamic world understanding AI agents.
The task naturally lacks training labels as the data collection and annotation procedure is extremely expensive.
As a remedy, leveraging the large-scale images is an attractive solution~\cite{dave2020tao}.
However, in doing so, we face three main issues: 1) lacking video supervision in images, 2) semantic label inconsistency between images and videos, and 3) the domain gap (e.g., explicit data styles or implicit data distributions are different) between images and videos.
The current learning paradigm bypasses the first two issues by independently training the detection head and tracking head with images and videos (\textit{decoupled}). 
Instead, our learning framework explicitly handles the former two issues by hallucinating the supervisions and enables end-to-end video model learning from all training data, leading to better feature representations (\textit{unified}).
The last issue is implicitly handled by the two-step training of image pre-training and video fine-tuning.
In the preliminary experiments, we observe a slight performance drop when concatenating image and video datasets as a single dataset, possibly due to the weaker feature adaptation toward the video domain.

The abovementioned issues are fundamental and compounded when jointly using the image and video labels for video recognition models.
We thus see they should be carefully considered and adequately handled.
Our proposal is an initial effort in this direction, and we believe more clever and innovative solutions will be developed and presented in the future.

\subsection{Data}
\label{sec:data}
\vspace{2mm}
\noindent{\textbf{LVIS}} \hspace{1mm}
LVIS~\cite{gupta2019lvis} is a large-scale benchmark for large vocabulary image recognition. 
It provides precise bounding boxes and masks annotations for various categories with the long-tailed distribution.
To be consistent with the prior works~\cite{dave2020tao,pang2021quasi}, we use LVIS v0.5 dataset and pre-train the model on 482 (out of 1230) LVIS categories that correspond to TAO categories.

\vspace{2mm}
\noindent{\textbf{TAO}} \hspace{1mm}
TAO~\cite{dave2020tao} is the first video benchmark for large vocabulary video recognition.
TAO dataset annotates 482 classes in total, which are the subset of LVIS dataset.
It has 400 videos, 216 classes in the training  set, 988 videos, 302 classes in the validation set, and 1419 videos, 369 classes in the test set.
The videos are annotated in 1 FPS.
We fine-tune the model on TAO-train and evaluate on TAO-val (or TAO-test).

\vspace{2mm}
\noindent We additionally use COCO~\cite{lin2014microsoft} and YTVIS~\cite{yang2019video} to evaluate the proposed teacher-student scheme on COCO $\rightarrow$ YTVIS transfer learning setup.

\vspace{2mm}
\noindent{\textbf{COCO 2017}} \hspace{1mm}
COCO contains 118k training images and 5k validation images.
We pre-train the model on 20 (out of 80) COCO categories, as the remaining 60 categories cannot be evaluated with YTVIS annotations.

\begin{figure}[t]
    \centering 
    \includegraphics[width=\textwidth]{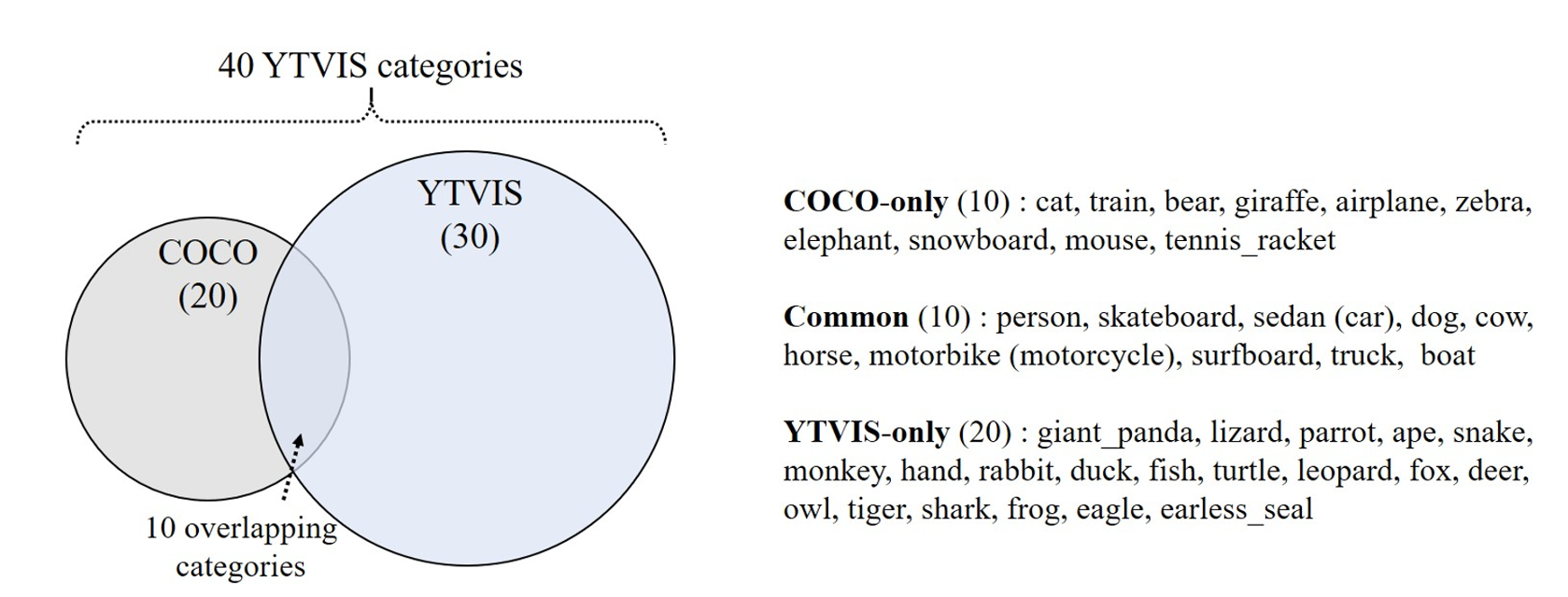}
    %\vspace{-1mm}
    \caption{
    We provide the detailed category distribution setup of COCO $\rightarrow$ YTVIS transfer learning used in the experiments.
    }
    \label{fig:coco_to_ytvis}
\end{figure}

\vspace{2mm}
\noindent{\textbf{YTVIS}} \hspace{1mm}
YTVIS is largest video benchmark for video instance segmentation.
YTVIS annotates 40 classes in total. It has 2238 training, 302 validation, 343 test video clips.
The videos are annotated in 5 FPS.
We fine-tune the model on 30 (out of 40) YTVIS categories to simulate missing 10 categories and new 20 categories during transfer learning (see~\figref{fig:coco_to_ytvis}).
We evaluate the model on YTVIS-val.

\subsection{Implementation Details}
\label{sec:imple}

\vspace{2mm}
\noindent{\textbf{Training.}} \hspace{1mm}
The proposals are implemented under the MMdetection framework~\cite{chen2019mmdetection}.
The COCO-style training schedule of 2$\times$ and 1$\times$ are adopted for LVIS pre-training and TAO fine-tuning, respectively.
We set the maximum number of predictions per image to 1000 for not losing correct predictions at frame-level~\cite{dave2021evaluating}, which makes the inter-frame object affinity matrix large and the subsequent tracking challenging. However, thanks to the proposed mosaic training, we see our tracker is robust to dense object tracking.
Batch size of 16 (2 per GPU) and an initial learning rate of 0.02 are used.
We randomly select a scale between 640 to 800 to resize the shorter side of images during training.
For the hyper-parameters of the models, we follow the original implementations~\cite{gupta2019lvis,wang2021seesaw,tan2021equalization,zhou2021probabilistic,pang2021quasi}.
The standard learning protocol is first pre-training the model in LVIS and then fine-tuning on TAO, \textit{i.e.}, decoupled learning.

\vspace{2mm}
\noindent{\textbf{Testing.}} \hspace{1mm}
Our method processes video frames recursively, generating detection boxes and matching them with the candidate tracks from the past frames.
Apart from the conventional tracking algorithms, we see the motion is highly irregular for tracking the large vocabulary of objects.
Thus, the most reliable way is to link detection boxes only based on their appearance features.
Here, the main matching strategy is a bi-directional softmax operation that examines the two matched objects being each other’s nearest neighbor in the embedding space~\cite{pang2021quasi}.
For the unmatched tracks, we keep them until it remains for more than 30 frames.
We use resized frames of 1080$\times$1080 for testing.

\vspace{2mm}
\noindent{\textbf{COCO $\rightarrow$ YTVIS transfer learning setup.}} \hspace{1mm}
We adopt MaskRCNN~\cite{he2017mask} with ResNet-50 FPN~\cite{he2016deep,lin2017feature} backbone for the COCO image pre-training.
We use 1$\times$ training pipeline.
We transfer the pre-trained MaskRCNN model weights to MaskTrack RCNN~\cite{yang2019video} for the YTVIS video fine-tuning.
At this stage, new weights are added to the class head, bounding box head, and mask head to accommodate newly added classes.
Also, the track head is appended to the model for object tracking.
We follow the original training schedules and hyper-parameters of MaskTrack RCNN~\cite{yang2019video}.

The presented two-step teacher-student scheme is applied to the model during transfer learning (see~\figref{fig:coco_to_ytvis}).
The distillation in the class head and the bounding box head follows the methods noted in the main paper.
For the mask head distillation, we collect teacher and student mask predictions and minimize their difference through MSE loss.

As the detailed class-wise evaluation in YTVIS is only possible for the 40 object categories, we simulate the pattern in~\figref{fig:coco_to_ytvis} by shrinking the original object categories in COCO and YTVIS.
In specific, for image pre-training, we trained the MaskRCNN on 20 COCO object categories.
For video fine-tuning, we trained the MaskTrack RCNN on YTVIS videos with 30 object categories, which consist of 10 overlapping object categories with the 20 COCO pre-trained categories and 20 new object categories.
The 10 overlapping object categories are selected based on the annotation frequency.
In this way, we can simulate the co-existence of missing object categories and new object categories during transfer learning.

\subsection{Oracle Analysis}
\label{sec:oracle}
To disentangle the impact of methods on object classification and tracking, we use two oracles: class oracle and track oracle on TAO validation set~\cite{dave2020tao}.

For \textbf{\textit{class oracle}}, we first compute the best matching between predicted and ground truth tracks in each video. 
The predicted tracks that match to a ground truth track with 3D IoU $>$ 0.5 are assigned the category of their matched ground truth track. 
Tracks that do not match to a ground truth track are treated as false positives.
This allows us to analyze the \textit{\textbf{pure tracking ability}} of models assuming the classification task is solved.

For \textit{\textbf{track oracle}}, we compute the best possible assignment of per-frame detection boxes to tracks, by comparing them with ground truth.
The class predictions for each detection are held constant. Any detection boxes that are not matched are removed. 
This allows us to evaluate the \textit{\textbf{pure classification ability}} of models given a perfectly linked per-frame detection boxes.

The results are summarized in~\tabref{tab:oracle}.
We use the same FasterRCNN-RFS tracker for the Decoupled and Unified methods.
We observe that our method outperforms the previous approaches in both oracle types.
This shows that the unified learning framework using all training data, LVIS and TAO, essentially improves the model's tracking and classification ability significantly.

\begin{table*}[t]
\centering
\resizebox{0.96\textwidth}{!}
{
\def\arraystretch{1.2}
\setlength{\tabcolsep}{16pt}
\begin{tabular}{l|ccr|ccr}
%\hlin/
\toprule
\multirow{2}{*}{Method}                    & \multicolumn{3}{c}{Oracle Class (pure tracking ability)}  & \multicolumn{3}{c}{Oracle Track (pure classification ability)} \\ \cline{2-7}
                                           &Track AP50 & Track AP75 & Track AP &Track AP50 & Track AP75 & Track AP \\
\hline
SORT~\cite{dave2020tao}                    & 30.2      & -          & -        & 31.5          & -              & -         \\
Decoupled~\cite{pang2021quasi}             & 34.7      & 12.2       & 15.1     &  32.1         & 12.1           & 14.8       \\
\textbf{Unified (Ours)}                    & \textbf{38.1} & \textbf{17.1}  & \textbf{18.4}&  \textbf{43.1} &  \textbf{16.7}   &  \textbf{19.9} \\
\bottomrule
\end{tabular}
}
\vspace{3mm}
\caption{
\textbf{Oracle analysis.}
We analyze the performance of two types of oracles: Oracle Class and Oracle Track.
The former provides the pure tracking ability of the model, and the latter allows us to analyze pure classification ability.
}
\label{tab:oracle}
\end{table*}

\bibliographystyle{splncs04}
\bibliography{egbib}
\end{document}